\documentclass[11pt,a4paper]{article}

\usepackage[utf8]{inputenc}
\usepackage[margin=1in]{geometry}
\usepackage{amsmath,amssymb}
\usepackage{graphicx}
\usepackage{booktabs}
\usepackage{array}
\usepackage{multirow}
\usepackage{xcolor}
\usepackage{hyperref}
\usepackage[numbers]{natbib}
\usepackage{newunicodechar}
\newunicodechar{，}{,}
\usepackage{caption}
\usepackage{subcaption}
\usepackage{float}
\usepackage{enumitem}
\usepackage{microtype}

\hypersetup{
    colorlinks=true,
    linkcolor=blue!70!black,
    citecolor=green!50!black,
    urlcolor=blue!70!black
}

\newcommand{\model}[1]{\textsc{#1}}
\newcommand{\ug}{$\mu$g/m$^3$}
\newcommand{\loio}{\textsc{loio}}

\definecolor{bestcolor}{RGB}{0,120,60}

\title{%
  \textbf{Evaluating the Generalizability of Foundation Models for Extreme Environmental Events: Case Study of California Wildfire PM2.5
}}

\usepackage{authblk}

\author[1]{Yongcan Huang}
\author[2,*]{Li Jiang}
\author[3]{Ze Yu Liu}

\affil[1]{\small College of Engineering, University of Georgia,
  Athens, Georgia, USA;
  Information System Department, UMBC, Baltimore, Maryland, USA ;
  \href{mailto:royhuang0421@gmail.com}{yhuang11@umbc.edu}}
\affil[2]{\small College of Graduate and Professional Studies,
  Trine University, Angola, Indiana, USA;
  \href{mailto:ljiang231@my.trine.edu}{ljiang231@my.trine.edu}}
\affil[3]{\small School of Professional Studies,
  Columbia University, New York, New York, USA;
  \href{mailto:zl3432@columbia.edu}{zl3432@columbia.edu}}

\affil[*]{\small Correspondence: ljiang231@my.trine.edu}

\date{\today}

\begin{document}
\maketitle

\begin{abstract}
Wildfire smoke events produce extreme PM$_{2.5}$ concentrations that pose
severe public health risks, yet accurately forecasting rare, hazardous-level
spikes remains a fundamental challenge. Time series foundation models
(TSFMs), large pretrained models offering zero-shot inference and
parameter-efficient adaptation, have shown strong performance on general
time-series benchmarks, but their behavior under extreme out-of-distribution
conditions is poorly understood. We present the first systematic benchmark
comparing six TSFM configurations (zero-shot TimesFM, Chronos-2, Moirai-2,
and Time-MoE, plus LoRA fine-tuned variants of Chronos-2 and Time-MoE)
against fully-trained deep learning baselines (LSTM, BiLSTM, Transformer) and
a na\"ive persistence reference on a 12-year (2013--2025) hourly PM$_{2.5}$
dataset covering 1,375 wildfire incidents across 79 California monitoring
sites. We design a leave-one-incident-out (LOIO) cross-validation protocol
that evaluates true generalization to unseen fire events, and assess models
using MAE, RMSE, and exceedance F1 at the U.S. Environmental Protection Agency (EPA) AQI severity thresholds across
forecast horizons of 6, 12, and 24 hours. Our results reveal a clear and
consistent hierarchy. The fully-trained BiLSTM achieves the lowest MAE
($5.16\,\mu g/m^3$) and the highest exceedance F1 at every AQI threshold,
including the Hazardous band ($>225.5\,\mu g/m^3$), where it reaches 0.63
against at most 0.54 for any foundation model. Zero-shot TSFMs improve on
na\"ive persistence in aggregate error, but the margin is modest, and
zero-shot Chronos-2 exhibits a severe RMSE tail instability
($23.4\,\mu g/m^3$, negative $R^2$) driven by sporadic large-magnitude
errors. LoRA fine-tuning on each fold's training incidents substantially
improves both adapted families and largely repairs the Chronos-2 instability,
but no foundation model, zero-shot or fine-tuned, surpasses the trained
recurrent baselines on any metric. These findings challenge the assumption
that larger pretrained models universally dominate in environmental
forecasting, and provide actionable deployment guidance for wildfire air
quality prediction.
\end{abstract}

\textbf{Keywords:} PM$_{2.5}$ forecasting; wildfire smoke; time series
foundation models; leave-one-incident-out; exceedance prediction; LoRA
fine-tuning; deep learning; air quality.

\section{Introduction}
\label{sec:intro}

Air pollution represents one of the most pressing environmental and public
health challenges of the 21st century. Among atmospheric pollutants,
particulate matter with an aerodynamic diameter of 2.5 micrometers or less
(PM$_{2.5}$) poses particularly serious health risks due to its ability to
penetrate deeply into the respiratory system and enter the bloodstream
\cite{who2021airquality}. Wildfires have emerged as an increasingly
significant source of PM$_{2.5}$, particularly under the prolonged drought
and rising temperatures associated with climate change
\cite{burke2021wildfire,reid2016critical}. Unlike urban and industrial
sources, wildfire smoke produces episodic but extreme concentration spikes
that persist for days and affect communities far downwind
\cite{jaffe2020wildfire}: in our dataset, hourly readings reach
1{,}249~\ug{}, roughly 35 times the Environmental Protection Agency (EPA) 24-hour standard of 35~\ug{}.
Accurate advance warning of hazardous concentrations is critical for public
health decisions including evacuation orders, school closures, and hospital
surge preparation, where the cost of a missed warning far outweighs the cost
of a false alarm.

Physics-based chemical transport models such as CMAQ and WRF-Chem provide
mechanistic representations of smoke transport but require substantial
computation, emission inventories, and meteorological inputs, limiting
real-time use \cite{byun2006cmaq,grell2005wrfchem,zhang2012ctmreview}.
Data-driven forecasting has therefore become the dominant alternative:
LSTM networks and their variants deliver strong gains over statistical
methods \cite{hochreiter1997lstm,li2017lstm_airquality,freeman2018lstmforecast},
and hybrid architectures combining convolution, recurrence, and attention now
anchor the PM$_{2.5}$ literature
\cite{vaswani2017attention,zeng2026iscience,liu2025hybridpm25,booth2025cross}.
For wildfire-specific PM$_{2.5}$, ensemble deep learning achieves strong
spatial prediction \cite{aguilera2021envhealth}, yet concentrations during
wildfires differ so sharply from routine urban emissions that models trained
on ambient data systematically underestimate maximum exposures
\cite{wu2025iop}. Despite this progress, existing studies share a critical
limitation: evaluations rely on chronological train--test splits or
single-site settings, neither of which assesses generalization to an unseen
wildfire incident.

A more recent paradigm shift comes from time series foundation models
(TSFMs), which are pretrained on large heterogeneous corpora and forecast
unseen series zero-shot, without task-specific training
\cite{das2024timesfm,ansari2024chronos,rasul2023lagllama}. Architecturally,
TSFMs span decoder-only attention (TimesFM), tokenization-based
language-model backbones (Chronos), masked-encoder universal forecasters
(Moirai), and sparse mixture-of-experts routing (Time-MoE, Moirai-MoE) in
which distinct experts can specialize to different temporal regimes; several
families additionally support parameter-efficient fine-tuning when limited
target data are available. Because pretraining promises to address the
cold-start problem that arises when a new fire ignites at a location with no
incident-specific history, TSFMs are conceptually attractive for wildfire air
quality. Yet despite their success in language and weather forecasting
\cite{fedus2022switch,lam2023graphcast}, TSFMs remain largely untested on air
quality, and whether their general forecasting ability extends to the
extreme, non-stationary dynamics of wildfire PM$_{2.5}$, and whether
lightweight fine-tuning meaningfully closes any gap, remains an open
question.

The present study addresses these gaps through a controlled benchmark of
trained deep learning baselines and TSFMs for wildfire PM$_{2.5}$
forecasting, evaluated under a leave-one-incident-out protocol that holds out
entire fire events. Our design examines three factors: (1)~\textbf{modeling
paradigm}: fully trained baselines (\model{BiLSTM}, \model{LSTM},
encoder-only Transformer) versus pretrained TSFMs, four zero-shot (TimesFM,
Chronos-2, Moirai-2, Time-MoE) and two LoRA fine-tuned (Chronos-2, Time-MoE)
configurations; (2)~\textbf{AQI severity stratum}: performance is evaluated
across five EPA 2024 exceedance thresholds (9.1,35.5, 55.5, 125.5, and
225.5~\ug) to distinguish behavior under moderate versus extreme
concentrations; and (3)~\textbf{forecast horizon}: 6, 12, and 24 hours ahead,
spanning near-term warning through next-day public health planning, since
accuracy typically degrades with lead time
\cite{zhang2012ctmreview,kumar2022forecasting}. All experiments share
identical preprocessing, training procedures, and evaluation metrics. We make
the following contributions:

\begin{enumerate}

\item \textbf{Benchmark and protocol.}
We release a 12-year, 79-site California wildfire PM$_{2.5}$
benchmark (1{,}375 incidents; 1.73M hourly records) together with a
leave-one-incident-out (LOIO) evaluation protocol that prevents
temporal leakage across fire events and directly simulates the
operationally critical scenario of predicting PM$_{2.5}$ for a
newly ignited fire with no prior incident-specific data.

\item \textbf{First TSFM benchmark for extreme wildfire PM$_{2.5}$.}
To our knowledge, this is the first systematic evaluation of
time series foundation models, across four zero-shot and two LoRA fine-tuned
configurations, against fully trained deep-learning baselines,
specifically targeting extreme, episodic wildfire PM$_{2.5}$
rather than routine urban pollution or LLM-based air quality pipelines.
All models are evaluated on the same univariate PM$_{2.5}$ input
to ensure a controlled and architecturally fair comparison across
paradigms.

\item \textbf{A counter-intuitive empirical finding.} A fully trained BiLSTM
consistently outperforms every TSFM configuration on both aggregate error and
extreme-event exceedance F1 across all forecast horizons. Specifically:
(i)~zero-shot TSFMs improve on the na\"ive persistence reference in aggregate
MAE, but only modestly (5.56--5.92 versus 6.44~$\mu g/m^3$), and none
approaches the trained recurrent baselines, indicating that large-scale
pretraining on general corpora transfers only weakly to the heavy-tailed,
non-stationary dynamics of wildfire smoke; (ii)~at the Hazardous threshold
($>225.5\,\mu g/m^3$), BiLSTM attains an exceedance F1 of 0.63 while zero-shot
TSFMs reach only 0.47--0.54, and zero-shot Chronos-2 additionally exhibits a
severe RMSE tail instability ($23.4\,\mu g/m^3$ with a negative $R^2$) that is
invisible to absolute-error metrics; (iii)~LoRA fine-tuning on each fold's
training incidents substantially improves both adapted families and largely
repairs the Chronos-2 instability, yet the fine-tuned variants still do not
surpass BiLSTM or LSTM on any metric, suggesting that lightweight in-domain
adaptation recalibrates magnitude scale but does not instill the fine-grained
temporal dynamics that govern extreme-event exceedance.

\item \textbf{Deployment guidance.}
We provide practical recommendations calibrated to data availability
and the target severity threshold, clarifying when and
whether zero-shot or LoRA fine-tuned TSFMs are operationally justified,
and identifying the conditions under which domain-specific
supervised training remains essential.

\end{enumerate}

The remainder of this paper is organized as follows.
Section~\ref{sec:related} reviews related work on PM$_{2.5}$ forecasting,
wildfire smoke modeling, and time series foundation models.
Section~\ref{sec:data} describes the dataset, preprocessing, and the
leave-one-incident-out protocol. Section~\ref{sec:methods} presents the
models, adaptation procedures, and evaluation metrics.
Section~\ref{sec:results} reports the results, and
Section~\ref{sec:conclusion} discusses the findings and future directions.
\section{Related Work}
\label{sec:related}

\subsection{PM$_{2.5}$ Forecasting with Deep Learning}
Recurrent and attention-based architectures now anchor data-driven air
quality forecasting. Long Short-Term Memory (LSTM) networks and their
gated and bidirectional variants remain strong baselines for pollutant
concentration prediction, and hybrid designs that pair convolutional or
autoencoder front-ends with recurrent or Transformer back-ends report
further gains by jointly modeling spatial structure and temporal
dependence. Recent examples span autoencoder and sparse-autoencoder
classifiers for next-hour NO\textsubscript{2}, PM\textsubscript{10} and
SO\textsubscript{2} in an industrial bay~\cite{rodriguezgarcia2025aealgeciras};
Bayesian-optimized CNN--LSTM--Transformer frameworks that fuse spatial
autocorrelation features for province-scale prediction in
Sichuan~\cite{zhang2025bocnnlstmtransformer}; systematic comparisons of
LSTM, CNN--LSTM, Transformer and Transformer--LSTM across forecasting
horizons~\cite{zeng2026iscience}; and Transformer--LSTM hybrids
tuned by metaheuristic search for regional PM\textsubscript{2.5}
prediction~\cite{liu2025hybridpm25}. A consistent finding is that
attention and frequency-aware mechanisms improve aggregate error metrics
(MAE, RMSE, $R^2$) and that explicit spectral or spatial encoding
mitigates the smoothing of sharp transitions characteristic of recurrent
models~\cite{leon2026multi}.

Despite this architectural diversity, the evaluation protocol in the
literature is remarkably uniform and conceals a gap directly relevant
to extreme-event forecasting. Models are almost always trained and
assessed on a chronological partition of a fixed set of
monitoring sites: the test segment is a temporal continuation of the same
stations seen in training, drawn from the same period and critically often
overlapping the same pollution episodes
~\cite{rodriguezgarcia2025aealgeciras,zhang2025bocnnlstmtransformer,
zeng2026iscience,liu2025hybridpm25}. Even benchmarks that test
spatial transfer do so within a single climatic region or province,
and even multi-site forecasting studies that apply formal significance
testing partition the data by time rather than by
event~\cite{leon2026multi}. The wildfire-specific literature is no
exception: graph-based forecasters of fire-influenced
PM\textsubscript{2.5} hold out whole years
(e.g., training on two seasons and testing on a third, while excluding an
anomalous low-activity year)~\cite{liao2025prescribedfiregnn}, and
LSTM-based fire early-warning systems train on one block of years and
test on the next~\cite{bhowmik2023multi}. These designs measure
interpolation within a known regime; they do not measure whether a model
generalizes to an unseen high-impact event whose dynamics were
absent from training.

This distinction matters most precisely where forecasting is hardest.
Wildfire smoke episodes are heavy-tailed, comparatively rare, and
mutually distinct in source location, transport, and intensity, so a
chronological split can leak information from a given fire's onset into
the training window used to predict its peak. To isolate genuine
event-level generalization, we adopt a leave-one-incident-out
cross-validation protocol in which each fold withholds the complete set
of records associated with one fire incident and evaluates transfer to
that held-out event. To our knowledge, this incident-holdout evaluation
has not been applied in prior PM\textsubscript{2.5} forecasting
benchmarks, and it is the basis on which we compare time-series
foundation models against trained recurrent and Transformer baselines in
the remainder of this work.

\subsection{Wildfire Smoke and PM$_{2.5}$ Modeling}
Wildfire-specific PM2.5 forecasting has been dominated by process-based models. These include coupled fire–atmosphere–chemistry systems such as WRF-Chem and WRF-SFIRE-CHEM, as well as operational smoke-forecasting systems such as NOAA's HRRR-Smoke ~\cite{ahmadov2017hrrrsmoke} and the HYSPLIT-based Smoke Forecasting System ~\cite{stein2015hysplit}, which produce 24–48 h forecasts from satellite fire detections, biomass-burning emission inventories, and meteorological fields ~\cite{kochanski2013wrfsc}. Such systems are mechanistically interpretable and operationally entrenched, yet they are computationally expensive, sensitive to uncertain emission and plume-rise assumptions, and tend to underestimate surface concentrations during the most severe episodes. Ensemble runs for the 2018 Camp Fire spanned nearly 1000~$\mu$g/m$^3$  across emission and plume-rise configurations ~\cite{li2020ensemble}, while a multi-system intercomparison for the 2019 Williams Flats fire revealed large inter-model spread and systematic bias ~\cite{ye2021williamsflats}. Tellingly, these assessments remain episodic case studies of individual named fires, underscoring both the difficulty of resolving extreme PM2.5 with process-based models and the absence of evaluation across many incidents under a common protocol.

Data-driven wildfire-smoke forecasting is comparatively recent and sparse.
Spatiotemporal deep learning has only begun to appear in this setting: a
spatiotemporal Transformer for hourly PM\textsubscript{2.5} in fire-prone
regions~\cite{yu2023predicting}, a graph neural network for
fire-influenced PM\textsubscript{2.5} across California that outperforms LSTM and
MLP baselines~\cite{liao2025prescribedfiregnn}, and a U-shaped LSTM early-warning
system anchored on the Camp Fire~\cite{bhowmik2023multi}. These studies
show that learned models can rival or exceed numerical guidance at short horizons
without explicit emission inventories. However, each is developed and validated
on a single region or a small number of named fires and partitions data chronologically or by year rather than
by event.

Crucially, neither paradigm has produced a standardized multi-incident
benchmark. Physics-based evaluations are typically episodic case studies of one
fire (e.g., the Camp, Williams Flats, or Caldor fires), and even multi-model
efforts are intercomparisons on a single event rather than one model
assessed across many events under a common protocol. Data-driven studies inherit
the same limitation, training and testing within a fixed region or season. As a
consequence, the field lacks a controlled assessment of how forecasters, either numerical
or learned, generalize across heterogeneous fire incidents that
differ in ignition, fuel load, terrain, and smoke transport. We address this gap
with a leave-one-incident-out benchmark spanning 1,375 California wildfire
incidents over 2013--2025, in which every held-out event is evaluated under a
common metric and horizon protocol.

\subsection{Time Series Foundation Models}
Transformer architectures entered time series analysis through models such as Informer and Autoformer, whose self-attention handled long-range dependencies and multivariate interactions more flexibly than recurrent or convolutional predecessors \cite{saravana2026transformers}. These models, however, were still trained from scratch per dataset. The foundation model era began in late 2023 with Lag-Llama \cite{rasul2023lagllama}, a decoder-only Transformer for univariate probabilistic forecasting, and TimesFM \cite{das2024timesfm}, a patch-based decoder-only model pre-trained on a large heterogeneous corpus. The 2024 wave broadened the design space: Chronos \cite{ansari2024chronos} tokenized series via scaling and quantization to treat forecasting as language modeling; Moirai \cite{woo2024moirai} introduced any-variate attention for universal multivariate forecasting; and Timer \cite{liu2024timer}, MOMENT \cite{goswami2024moment}, and ChatTime \cite{wang2025chattime} added autoregressive, encoder-based, and multimodal variants, respectively. Broader taxonomies and systematic reviews trace this rapid expansion across hundreds of studies \cite{abdullahi2025llm}\cite{liang2024kdd}\cite{liu2026survey}\cite{ma2024survey}.

The 2025 frontier advanced along two axes: scale and second-generation refinement. Time-MoE \cite{shi2025timemoe} reached 2.4 billion parameters via a sparse Mixture-of-Experts architecture, empirically confirming that scaling laws hold for temporal data, while Sundial \cite{liu2025sundial} enabled native probabilistic pre-training through a flow-matching TimeFlow Loss. The two most widely deployed families also advanced to second-generation releases: Chronos-2 \cite{ansari2025chronos2}, a 120M-parameter encoder-only model whose group-attention mechanism extends zero-shot forecasting to multivariate and covariate-informed tasks, and Moirai-2 \cite{liu2025moirai2}, which replaces the masked encoder with a decoder-only backbone trained via quantile forecasting and multi-token prediction. In parallel, domain-focused work has begun to question the generality of these models: \cite{cohen2026time} show with the BOOM benchmark that general-purpose TSFMs underperform on heavy-tailed, nonstationary observability metrics, motivating domain-specialized evaluation.

\section{Dataset and Preprocessing}
\label{sec:data}

\subsection{Data Sources}
We compiled an event-aligned air-quality panel of hourly PM2.5 measurements drawn from 79 EPA-certified monitoring stations across California, spanning February 2013 through November 2025. Wildfire incident metadata including ignition location, creation and containment dates, and final burned area, were obtained from the CAL FIRE incident archive, and the corresponding PM2.5 observations were retrieved from the California Air Resources Board (CARB) AQMIS interface. To align fire activity with the downwind air-quality response, each incident was paired with its nearest monitoring station, defined as the EPA-certified site that minimizes the geodesic (great-circle) distance between the station coordinates and the incident location. Incidents whose nearest station exceeded 100 km were excluded, as measurements beyond this range are unlikely to capture the fire's smoke signal. A single station may serve as the nearest site for multiple incidents, since fires recurring in the same region share the same downwind monitor.

\begin{figure}[!htbp]
\centering
\includegraphics[width=1\linewidth]{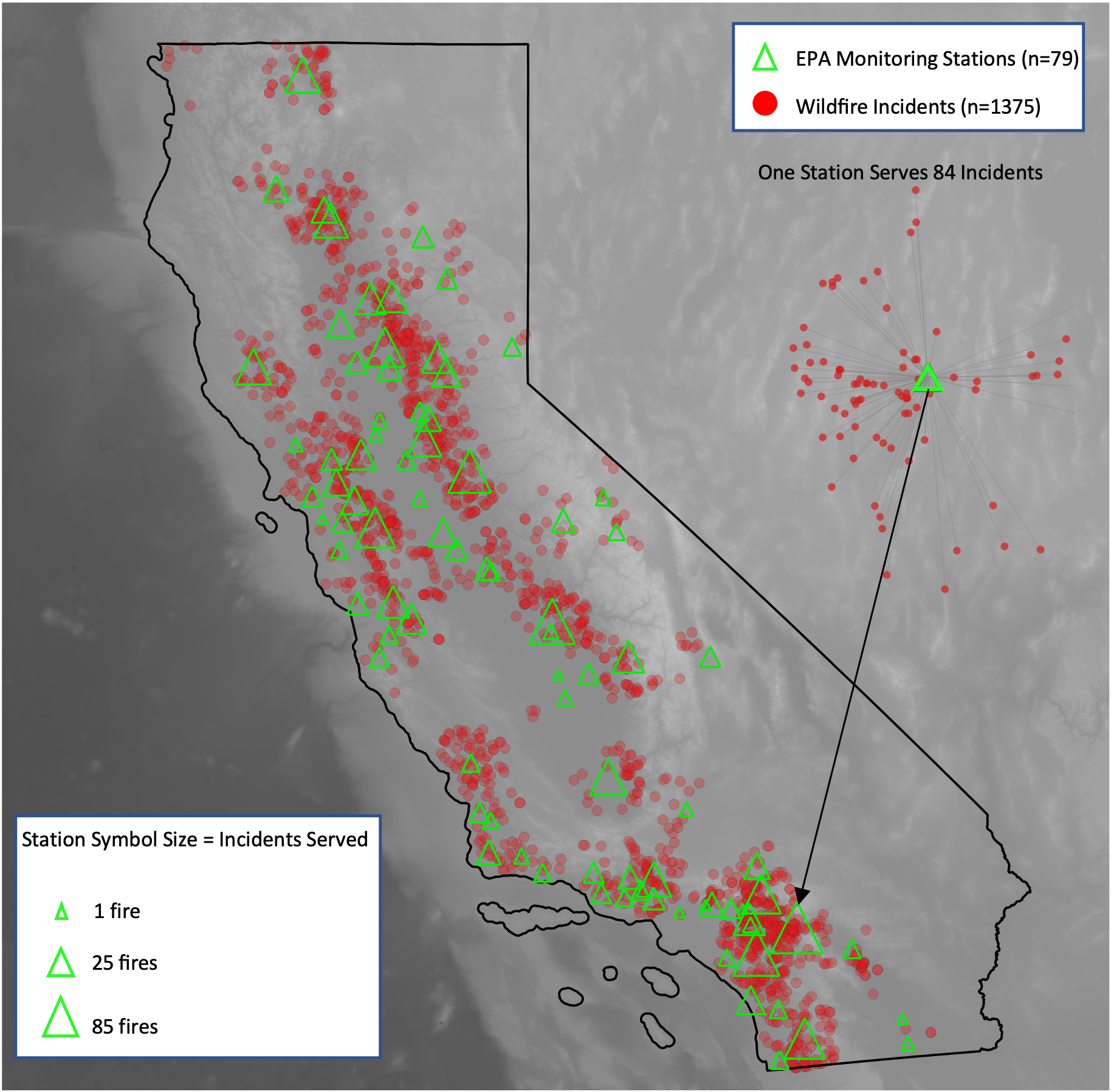}
\caption{Spatial distribution of the 79 EPA-certified PM$_{2.5}$ monitoring
stations (green triangles) and the 1{,}375 wildfire incidents (red points)
included in the study, overlaid on shaded-relief terrain. Each station is
paired with the wildfire incidents for which it is the nearest monitor; the
size of each station symbol is scaled by the number of incidents it serves.
The inset illustrates one representative station that serves 84 incidents,
highlighting the one-to-many wildfire--station pairing.}
\label{fig:study_area}
\end{figure}

Together with each incident's creation and containment dates, these wildfire–station pairs defined the retrieval keys for data collection. Rather than downloading a single static archive, we assembled the panel with an incident-aware crawler that queried each wildfire–station pair over its corresponding incident window. Multi-year windows were decomposed into calendar-year requests, and downloads ran asynchronously under bounded concurrency, with retry logic and file-level deduplication to accommodate the length of the historical record and the query limits of the remote interface. Figure~\ref{fig:study_area} shows the spatial distribution of the 79 monitoring stations and the associated wildfire events, illustrating both the statewide coverage of the sensor network and the geographic spread of the incidents analyzed in this study.

\subsection{Preprocessing}
\label{sec:preprocessing}

The raw CARB exports required cleaning and harmonization before modeling.
Non-tabular metadata blocks were removed, and the cleaned records were
consolidated into an hourly panel indexed by monitoring site, timestamp, and
wildfire incident. Because a single monitor is frequently the nearest station
for several nearby fires, a given site--hour measurement can be associated with
more than one incident window; the underlying observations were therefore
de-duplicated to 650,091 unique site--hour PM\textsubscript{2.5} measurements,
which expand to 1,726,019 incident-aligned records. 

As hourly monitoring records contain gaps from sensor downtime and
transmission loss, we deliberately avoided interpolation, which would inject
artificial dynamics into a series whose extreme events are the object of study.
Instead, each site's record was partitioned into internally continuous segments
at temporal breaks exceeding two hours, and all subsequent windowing was
confined within these segments so that no input or forecast window spans a gap.
Negative readings (2.80\% of unique measurements), attributable to instrument
noise near the low-concentration detection limit, were clipped to zero.
Predictors were standardized using per-site $z$-score normalization, which
preserves each station's physical baseline while rendering heterogeneous series
comparable, whereas forecast targets were retained in native
$\mu$g/m\textsuperscript{3} units to keep evaluation metrics interpretable
against regulatory AQI thresholds. Modeling used the univariate
PM\textsubscript{2.5} series.

Supervised examples were then constructed with a sliding window comprising a
48-hour input context and three forecast horizons of 6, 12, and 24 hours,
advanced with a stride of 6 hours; the three horizons share a common set of
input windows so that their results are directly comparable. Incident sequences
shorter than 72 hours which is the minimum span required to form a single input--target
pair at the longest horizon were excluded, retaining 891 of the 1,375
sequences and yielding 277,866 windowed examples. To probe performance under the
conditions of greatest operational concern rather than on routine background air
quality, evaluation was conducted on held-out continuous weeks containing
elevated-concentration episodes.

\begin{figure}[!htbp]
\centering
\includegraphics[width=\linewidth]{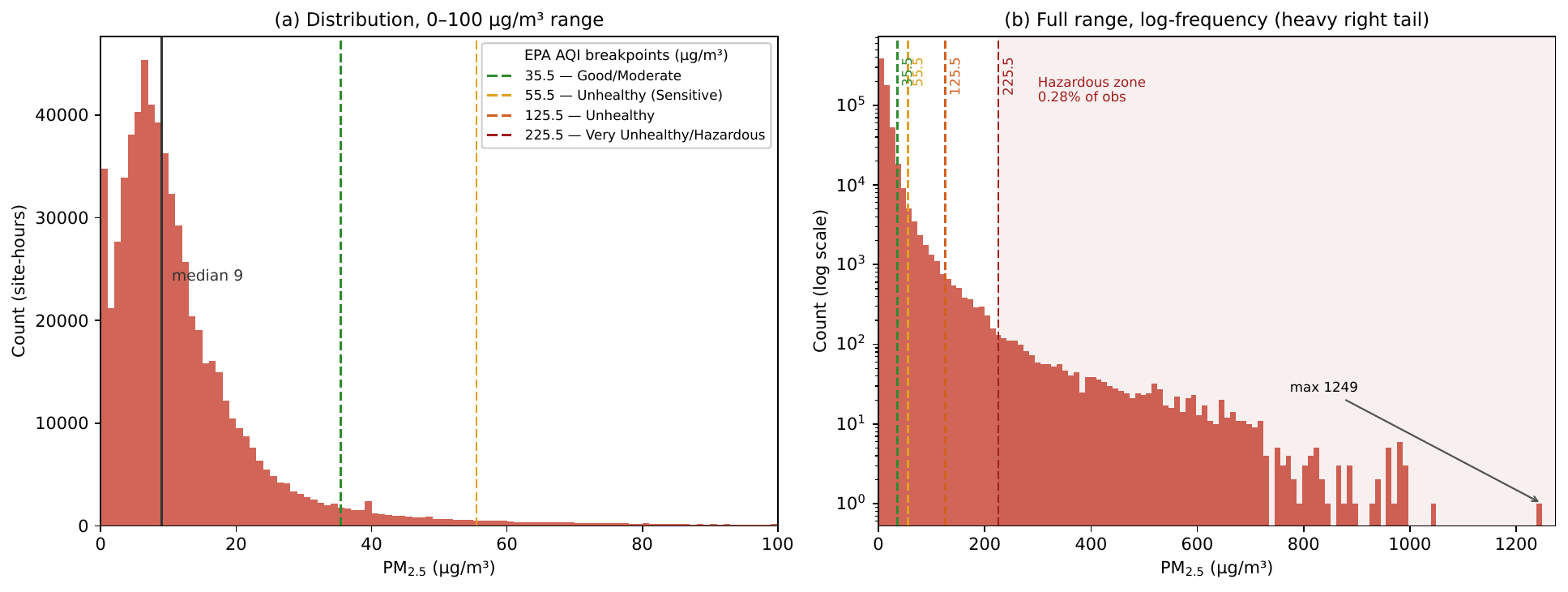}
\caption{Distribution of hourly PM\textsubscript{2.5} concentrations across the
79 California monitoring sites (650,091 unique site--hour measurements).
(a)~The 0--100~$\mu$g/m\textsuperscript{3} range (median
9~$\mu$g/m\textsuperscript{3}); dashed lines mark the EPA AQI breakpoints at
35.5, 55.5, 125.5, and 225.5~$\mu$g/m\textsuperscript{3}, separating the
Good/Moderate, Unhealthy for Sensitive Groups, Unhealthy, and Very
Unhealthy/Hazardous categories, respectively. (b)~The full range on a
logarithmic frequency scale, revealing the heavy right tail: extreme
concentrations extend to 1,249~$\mu$g/m\textsuperscript{3}, with only 0.28\% of
observations reaching Hazardous levels
($>$225.5~$\mu$g/m\textsuperscript{3}). }
\label{fig:pm25dist}
\end{figure}

Table~\ref{tab:dataset} and Figure~\ref{fig:pm25dist} summarize the resulting
distribution, which is severely right-skewed (skewness $\approx 12$). The median
hourly concentration is only 9~$\mu$g/m\textsuperscript{3}, and 93.8\% of
measurements fall within the Good-to-Moderate range (below
35.5~$\mu$g/m\textsuperscript{3}), yet the upper tail extends to a maximum of
1,249~$\mu$g/m\textsuperscript{3}, with 0.28\% of observations reaching Hazardous
levels (above 225.5~$\mu$g/m\textsuperscript{3}). This pronounced imbalance---
abundant background hours against sparse but consequential extremes---directly
motivates the extreme-event evaluation protocol adopted in this study.

\begin{table}[t]
\centering
\caption{Dataset summary statistics.}
\label{tab:dataset}
\begin{tabular}{@{}ll@{}}
\toprule
Attribute & Value \\
\midrule
Time span                         & February 2013 -- November 2025 (hourly) \\
Unique site--hour measurements    & 650,091 \\
Incident-aligned records          & 1,726,019 \\
Retained sequences                & 891 of 1,375 ($\geq$72\,h) \\
Windowed examples                 & 277,866 \\
Monitoring sites                  & 79 EPA sites across California \\
Input / forecast horizons         & 48\,h input; 6, 12, 24\,h ahead (stride 6\,h) \\
PM\textsubscript{2.5} range       & 0 -- 1,249 $\mu$g/m\textsuperscript{3} (negatives clipped to 0) \\
Median / Mean PM\textsubscript{2.5} & 9 / 14 $\mu$g/m\textsuperscript{3} \\
Good/Moderate ($<$35.5)           & 93.8\% \\
Hazardous ($>$225.5)              & 1,832 (0.28\%) \\
Negative readings (clipped)       & 2.80\% \\
Missing values                    & None \\
\bottomrule
\end{tabular}
\end{table}

\subsection{Leave-One-Incident-Out Protocol}
\label{sec:loio}
Standard temporal splits are poorly suited to wildfire forecasting because a
single fire can persist for days or months, producing many overlapping windows
with shared meteorology, source emissions, and site-specific background
conditions. A naive chronological split can therefore place early windows from
one fire in training and later windows from the same fire in testing, yielding
optimistic performance estimates through incident-level leakage.

To avoid this failure mode, we adopt a grouped leave-one-incident-out
(\loio{}) cross-validation protocol. The panel first defines one hourly sequence
for each usable \((\texttt{incident\_id}, \texttt{site\_id})\) pair. Sequences
shorter than the minimum length required for the full forecasting task are
excluded: with an input window of $L_{\text{in}}=48$ hours and a maximum forecast
horizon of $H_{\max}=24$ hours, each retained sequence must contain at least
\(L_{\text{in}}+H_{\max}=72\) consecutive hourly records. This filtering reduces
the 1{,}375 incident--site sequences to 891 usable sequences spanning 79
monitoring sites, each with no internal hourly gaps. For a retained sequence of
length \(L\), the number of sliding windows generated at stride \(s=6\) hours is
\begin{equation}
N_w(L) \;=\; \left\lfloor \frac{L - (L_{\text{in}} + H_{\max})}{s} \right\rfloor + 1 ,
\label{eq:nwindows}
\end{equation}
and the three horizons (6, 12, and 24 hours) share this common set of input
windows so that their results are directly comparable. Crucially, sliding
windows are generated only \emph{after} fold assignment, so all windows derived
from the same sequence inherit the same fold label and never straddle the
train--test boundary.

The fold construction is severity-stratified. For each retained sequence \(i\)
we compute its peak concentration
\(p_i = \max_{t} \text{PM}_{2.5}^{(i)}(t)\) and assign it to a severity quintile
\(Q(i)\in\{1,\dots,5\}\) based on the empirical quintiles of \(\{p_i\}\). Within
each quintile the sequences are shuffled with a fixed random seed and
distributed round-robin across the \(K=5\) folds:
\begin{equation}
\text{fold}(i) \;=\; \pi_{Q(i)}(i) \bmod K ,
\label{eq:foldassign}
\end{equation}
where \(\pi_{Q(i)}\) denotes the seeded permutation of sequences within quintile
\(Q(i)\). This yields folds balanced not only in sequence count, but also in the
long-tailed smoke-severity distribution that dominates wildfire forecasting
difficulty. In each evaluation run, all windows from the held-out fold are
excluded from training, ensuring that no test window has a training counterpart
from the same wildfire--site episode. The resulting benchmark contains
277{,}866 windows, each using 48 hours of PM$_{2.5}$ history to predict
PM$_{2.5}$ over 6-, 12-, and 24-hour horizons.

Table~\ref{tab:folds} reports the resulting fold composition. The mean peak
PM$_{2.5}$ values remain comparable across folds despite extreme events reaching
nearly 1{,}250~\ug{}, and the 24-hour exceedance rates fall within a narrow
range. This balance matters because rare high-smoke episodes drive much of the
operational forecasting challenge, yet are precisely the cases most easily
misrepresented by ungrouped temporal or random splits.

\begin{table}[H]
\centering
\caption{Five-fold \loio{} cross-validation statistics.}
\label{tab:folds}
\small
\begin{tabular}{ccccc}
\toprule
\textbf{Fold} & \textbf{Incidents} & \textbf{Windows} &
\textbf{Mean max PM$_{2.5}$ (\ug)} & \textbf{Exceedance rate ($>$35.5)} \\
\midrule
0 & 180 & 61{,}277 & 146.9 & 21.3\% \\
1 & 180 & 56{,}420 & 138.3 & 21.0\% \\
2 & 179 & 56{,}204 & 131.7 & 22.7\% \\
3 & 176 & 50{,}854 & 128.5 & 18.0\% \\
4 & 176 & 53{,}111 & 144.7 & 19.3\% \\
\midrule
\textbf{Total} & \textbf{891} & \textbf{277{,}866} & 138.0 & 20.5\% \\
\bottomrule
\end{tabular}
\end{table}

\begin{figure}[!htbp]
\centering
\includegraphics[width=\linewidth]{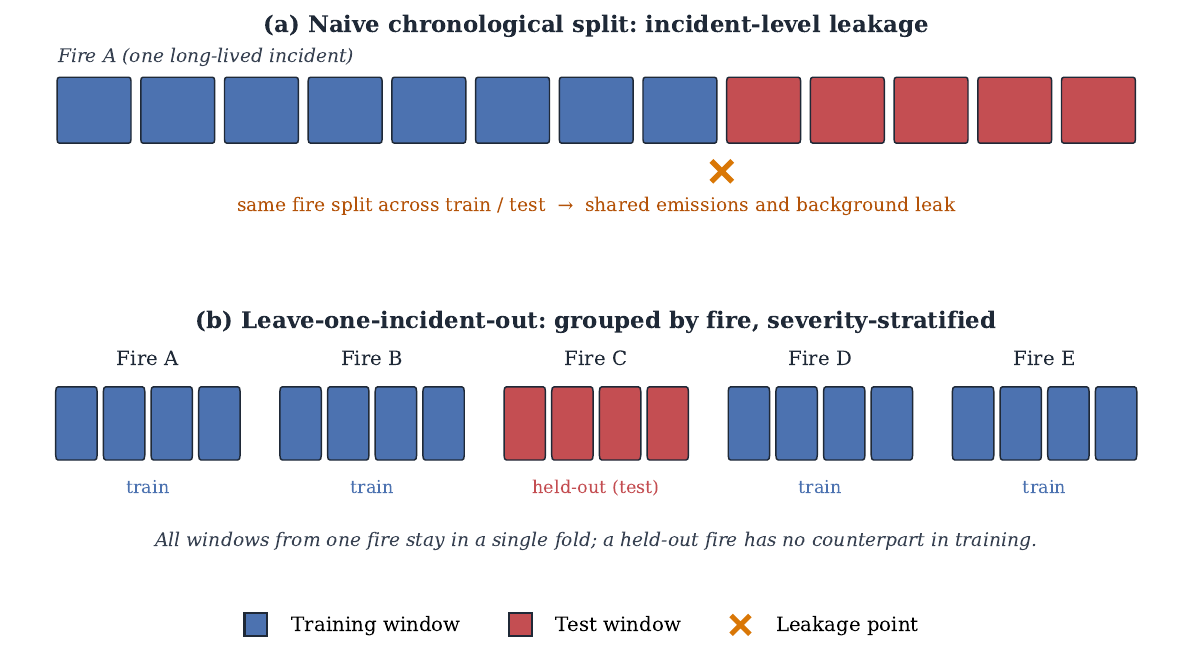}
\caption{Comparison of data-splitting strategies. (a)~A naive chronological
split places early windows of a long-lived fire in training and later windows of
the same fire in testing, leaking shared emissions, meteorology, and
site background across the train--test boundary. (b)~The proposed
leave-one-incident-out protocol groups all windows from a given
wildfire--site episode into a single fold, so that a held-out fire has no
counterpart in training. Folds are additionally stratified by smoke severity.}
\label{fig:loio}
\end{figure}

\section{Methods}
\label{sec:methods}

We frame wildfire-driven PM$_{2.5}$ forecasting as a univariate multi-horizon
regression problem and benchmark a non-learning persistence reference together
with three families of learned models under the identical
leave-one-incident-out protocol of Section~\ref{sec:loio}: a set of trained
neural baselines, a set of pretrained time-series foundation models (TSFMs)
evaluated zero-shot, and LoRA-adapted variants of two of these foundation models.
This section defines the forecasting task, then details each model and how its
inputs and outputs are handled.

\subsection{Input Representation}
\label{sec:input}

All models receive the same underlying supervised examples: a 48-hour history
of hourly PM$_{2.5}$ used to forecast the subsequent 6, 12, and 24 hours, but
the numerical form in which that history is presented differs between the
trained baselines and the pretrained foundation models, reflecting how each
class of model expects its inputs to be scaled.

\paragraph{Common windowed tensor.}
Preprocessing produces a single tensor
$\mathbf{X}\in\mathbb{R}^{N\times L_{\text{in}}\times F}$ with $N=277{,}866$
windows, input length $L_{\text{in}}=48$, and $F$ stored channels, together with
horizon-specific targets $\mathbf{y}^{(h)}\in\mathbb{R}^{N\times h}$ for
$h\in\{6,12,24\}$ retained in native \ug{}. The input
for window $i$ is the scalar sequence $\mathbf{x}_i=(x_{i,1},\dots,x_{i,48})$.

\paragraph{Per-site standardization.}
The stored history is standardized per monitoring site. For a window belonging
to site $s$, each value is transformed as
\begin{equation}
\tilde{x}_{i,t} \;=\; \frac{x_{i,t}-\mu_s}{\sigma_s},
\label{eq:zscore}
\end{equation}
where $\mu_s$ and $\sigma_s$ are the mean and standard deviation of PM$_{2.5}$ at
site $s$, estimated on training data. This standardized form
$\tilde{\mathbf{x}}_i$ is the representation consumed directly by the trained
baselines; the site statistics $(\mu_s,\sigma_s)$ are retained so that any
prediction can be mapped back to physical units.

\paragraph{Trained baselines (LSTM, BiLSTM, Transformer).}
These models ingest the standardized sequence
$\tilde{\mathbf{x}}_i\in\mathbb{R}^{48}$ as a length-48, single-channel input.
To decouple learning from each site's baseline level, targets are expressed as
residuals relative to the last observed (standardized) value
$\tilde{x}_{i,48}$; the model predicts $\Delta^{(h)}_i$ such that the
standardized forecast is $\tilde{x}_{i,48}+\Delta^{(h)}_i$, which is then
de-standardized for evaluation. This residual parameterization provides a strong
persistence-anchored starting point and stabilizes training under the
heavy-tailed concentration distribution.

\paragraph{Pretrained foundation models (Chronos-2, Moirai-2, TimesFM, Time-MoE).}
Zero-shot time-series foundation models apply their own internal instance
normalization and therefore expect inputs on the native measurement scale.
Accordingly, the standardized history is inverted back to \ug{} via
$x_{i,t}=\sigma_s\tilde{x}_{i,t}+\mu_s$ before being supplied as the context
sequence. Each backbone then tokenizes this context according to its own scheme:
TimesFM applies continuous-value quantile forecasting with input normalization
enabled, Chronos-2 uses learned value tokenization, Moirai-2 relies on
patch-based encoding, and Time-MoE performs autoregressive point generation. All
models are queried through their public pretrained interfaces without any
gradient updates. Time-MoE is the exception among the foundation models: it is
operated on the standardized sequence $\tilde{\mathbf{x}}_i$, consistent with its
pretraining convention, and its outputs are de-standardized for evaluation. In
every case the point forecast is taken as the predictive median, and all metrics
are computed in native \ug{} so that results are directly comparable across
model classes and against regulatory AQI thresholds.
Table~\ref{tab:input} summarizes how the input representation differs across the
model classes.

\begin{table}[t]
\centering
\caption{Input representation across model classes. All models are univariate
(PM$_{2.5}$ only) and share the 48-hour context and 6/12/24-hour horizons; they
differ in input scale, target parameterization, and forecasting mechanism.
Enc.\ = encoder-only; dec.\ = decoder-only; MoE = mixture-of-experts.}
\label{tab:input}
\resizebox{\textwidth}{!}{%
\begin{tabular}{@{}lllll@{}}
\toprule
\textbf{Model} & \textbf{Type} & \textbf{Input scale} &
\textbf{Target} & \textbf{Mechanism} \\
\midrule
Na\"ive Persistence & Reference & Native \ug{} & Last-value carry-forward & None \\
\midrule
LSTM        & Trained   & Per-site $z$-score & Residual from last value & Recurrent \\
BiLSTM      & Trained   & Per-site $z$-score & Residual from last value & Bidirectional recurrent \\
Transformer & Trained   & Per-site $z$-score & Residual from last value & Self-attention \\
\midrule
Chronos-2   & Zero-shot & Native \ug{} & Direct (median quantile) & Patch; enc., quantile head \\
Moirai-2    & Zero-shot & Native \ug{} & Direct (median quantile) & Patch; dec., multi-token quantile \\
TimesFM     & Zero-shot & Native \ug{} & Direct (median quantile) & Patch; dec., quantile head \\
Time-MoE    & Zero-shot & Per-site $z$-score & Direct (autoregressive) & Point-wise; dec., MoE \\
\midrule
Chronos-2   & LoRA-FT & Native \ug{} & Direct (median quantile) & Patch; enc., quantile head \\
Time-MoE    & LoRA-FT & Per-site $z$-score & Direct (autoregressive) & Point-wise; dec., MoE \\
\bottomrule
\end{tabular}%
} 
\end{table}

\subsection{Problem Formulation}
\label{sec:problem}

For a window $i$ let $\mathbf{x}_i=(x_{i,1},\dots,x_{i,L_{\text{in}}})$ denote the
observed hourly PM$_{2.5}$ history with $L_{\text{in}}=48$, and let
$\mathbf{y}_i^{(h)}=(y_{i,1},\dots,y_{i,h})$ denote the subsequent $h$ hourly
values for horizon $h\in\mathcal{H}=\{6,12,24\}$. A forecaster is a function
\begin{equation}
f_\theta:\ \mathbb{R}^{L_{\text{in}}}\ \rightarrow\ \mathbb{R}^{h},
\qquad
\hat{\mathbf{y}}_i^{(h)} = f_\theta\!\left(\mathbf{x}_i;\,h\right),
\label{eq:task}
\end{equation}
producing a point forecast for each horizon. All targets are evaluated in native
\ug{}. The three horizons are produced from a common 48-hour context so that
results are directly comparable across horizons and models.

\subsection{Na\"ive Persistence Reference}
\label{sec:naive}

As a non-learning lower bound we include a na\"ive persistence forecaster, which
propagates the last observed concentration across the entire horizon,
\begin{equation}
\hat{y}^{(h)}_{i,\tau} = y_{i,L_{\text{in}}}, \qquad \tau=1,\dots,h .
\label{eq:persistence}
\end{equation}
Persistence carries no trainable parameters and is applied directly in native
\ug{}. Because hourly PM$_{2.5}$ is strongly autocorrelated, it is a deliberately
demanding reference at short horizons and defines the skill floor that any
learned model must exceed to demonstrate genuine predictive value. It is also the
implicit anchor of the trained baselines, whose residual parameterization
(Eq.~\ref{eq:residual}) predicts departures from exactly this last-value
forecast; reporting persistence explicitly therefore isolates the contribution of
the learned residual from the autocorrelation the models inherit for free.

\subsection{Trained Neural Baselines}
\label{sec:baselines}

The trained baselines consume the per-site standardized history
$\tilde{\mathbf{x}}_i$ (Eq.~\ref{eq:zscore}) and predict a residual relative to
the last observed standardized value $\tilde{x}_{i,L_{\text{in}}}$. Writing the
standardized target as
$\tilde{y}^{(h)}_{i,\tau}=(y^{(h)}_{i,\tau}-\mu_s)/\sigma_s$, each model is
trained to output
\begin{equation}
\Delta^{(h)}_{i,\tau} \;=\; \tilde{y}^{(h)}_{i,\tau} - \tilde{x}_{i,L_{\text{in}}},
\label{eq:residual}
\end{equation}
and predictions are reconstructed to physical units by inverting the
standardization and clipping at zero,
\begin{equation}
\hat{y}^{(h)}_{i,\tau}
= \max\!\Big(0,\ \big(\hat{\Delta}^{(h)}_{i,\tau}+\tilde{x}_{i,L_{\text{in}}}\big)\,\sigma_s+\mu_s\Big).
\label{eq:reconstruct}
\end{equation}
This residual parameterization anchors each forecast to persistence and
stabilizes learning under the heavy-tailed concentration distribution. All three
baselines share a common backbone width of $d=64$, two layers, dropout $0.1$,
and a separate linear head per horizon; they are trained by minimizing mean
squared error on $\Delta^{(h)}$ with early stopping.

\paragraph{LSTM.}
A two-layer unidirectional LSTM~\citep{hochreiter1997lstm} encodes the sequence.
At each step the standard recurrence updates the hidden and cell states,
\begin{equation}
(\mathbf{h}_t,\mathbf{c}_t)=\mathrm{LSTM}(\tilde{x}_{i,t},\mathbf{h}_{t-1},\mathbf{c}_{t-1}),
\qquad \mathbf{h}_t\in\mathbb{R}^{d},
\label{eq:lstm}
\end{equation}
and the final hidden state $\mathbf{h}_{L_{\text{in}}}$ is mapped to each horizon
by a linear head,
$\hat{\boldsymbol{\Delta}}^{(h)}_i = \mathbf{W}_h\mathbf{h}_{L_{\text{in}}}+\mathbf{b}_h$.

\paragraph{BiLSTM.}
A bidirectional variant runs two LSTMs over the context in forward and backward
directions, yielding hidden states $\overrightarrow{\mathbf{h}}_t$ and
$\overleftarrow{\mathbf{h}}_t$. The final representation concatenates the two
terminal states,
\begin{equation}
\mathbf{h}^{\text{bi}} = \big[\,\overrightarrow{\mathbf{h}}_{L_{\text{in}}}\ ;\ \overleftarrow{\mathbf{h}}_{1}\,\big]\in\mathbb{R}^{2d},
\label{eq:bilstm}
\end{equation}
which is passed to per-horizon linear heads. Bidirectional context lets the
encoder summarize the full 48-hour window symmetrically, at the cost of doubling
the representation width.

\paragraph{Transformer.}
The Transformer baseline~\citep{vaswani2017attention} projects each scalar input
to width $d$, adds sinusoidal positional encodings, and applies a two-layer
encoder with $4$ attention heads and a feed-forward width of $4d$. For a
projected, position-encoded sequence
$\mathbf{Z}\in\mathbb{R}^{L_{\text{in}}\times d}$, each layer computes multi-head
self-attention,
\begin{equation}
\mathrm{Attn}(\mathbf{Q},\mathbf{K},\mathbf{V})=\mathrm{softmax}\!\left(\frac{\mathbf{Q}\mathbf{K}^{\top}}{\sqrt{d_k}}\right)\mathbf{V},
\label{eq:attn}
\end{equation}
with $\mathbf{Q},\mathbf{K},\mathbf{V}$ linear projections of $\mathbf{Z}$ and
$d_k=d/4$. The pooled encoder output feeds per-horizon linear heads.
Self-attention provides direct access to all lags in the window without
recurrence.

\subsection{Pretrained Foundation Models}
\label{sec:tsfm}

The foundation models are evaluated zero-shot: their pretrained weights
are frozen and queried through their public forecasting interfaces, with no
gradient updates on wildfire data. Except where noted, each model applies its
own internal instance normalization and therefore receives the history on the
native \ug{} scale, obtained by inverting Eq.~\eqref{eq:zscore}. The
probabilistic backbones (Chronos-2, Moirai-2, and TimesFM) map a context
sequence to a predictive distribution over the horizon,
\begin{equation}
p_\phi\!\left(\mathbf{y}^{(h)}\mid \mathbf{x}\right),
\qquad
\hat{\mathbf{y}}^{(h)} = \mathrm{median}\big[p_\phi(\cdot\mid\mathbf{x})\big],
\label{eq:tsfm}
\end{equation}
from which the point forecast is taken as the predictive median
($q = 0.5$); Time-MoE instead emits point forecasts directly through
autoregressive generation. The models differ in how the context is tokenized
and how the forecast is parameterized, as detailed below.

\paragraph{Chronos-2.}
Chronos-2~\citep{ansari2025chronos2} is a 120M-parameter, encoder-only
Transformer that generalizes its predecessor's language-modeling formulation:
rather than autoregressively generating tokens from a discrete value
vocabulary, it embeds the scaled context as patches, attends over them with a
group-attention mechanism that also enables multivariate and covariate-informed
forecasting, and emits multi-step quantile forecasts for the horizon in a
single forward pass. We use its univariate interface and take the median
quantile ($q=0.5$) as the point prediction.

\paragraph{Moirai-2.}
Moirai-2~\citep{liu2025moirai2} is a patch-based encoder. The context is divided
into temporal patches that are embedded as tokens; a Transformer encoder attends
over these patch tokens and emits a quantile forecast over the horizon. We use a
context length of $48$, a single target dimension, and the median quantile as the
point prediction.

\paragraph{TimesFM.}
TimesFM~\citep{das2024timesfm} segments the context into non-overlapping patches,
each embedded through a residual MLP into a token, and a decoder-only Transformer
predicts subsequent patches. We enable input normalization and a continuous
quantile head with positivity and quantile-crossing constraints, using the median
as the point forecast.

\paragraph{Time-MoE.}
Time-MoE~\citep{shi2025timemoe} is a decoder-only, point-wise mixture-of-experts
forecaster that generates the horizon autoregressively, one step at a time.
Consistent with its pretraining convention, it is the one foundation model
operated on the per-site standardized sequence $\tilde{\mathbf{x}}_i$; its
autoregressive outputs are de-standardized through the inverse of
Eq.~\eqref{eq:zscore} for evaluation.

\subsection{LoRA Fine-Tuning}
\label{sec:fewshot}

To quantify how much a small amount of in-domain adaptation narrows the gap
between zero-shot foundation models and the trained baselines, we additionally
evaluate parameter-efficient fine-tuned variants of two foundation models,
Chronos-2 and Time-MoE. Rather than updating the full backbone, we attach
low-rank adaptation (LoRA) modules~\citep{hu2022lora} to the pretrained weights,
so that only a small set of adapter parameters is trained while the original
weights remain frozen. For a frozen weight matrix
$\mathbf{W}_0\in\mathbb{R}^{d_\text{out}\times d_\text{in}}$, LoRA
reparameterizes the update as a low-rank product,
\begin{equation}
\mathbf{W} = \mathbf{W}_0 + \frac{\alpha}{r}\,\mathbf{B}\mathbf{A},
\qquad
\mathbf{B}\in\mathbb{R}^{d_\text{out}\times r},\ \mathbf{A}\in\mathbb{R}^{r\times d_\text{in}},\ r\ll d,
\label{eq:lora}
\end{equation}
where $r$ is the adapter rank and $\alpha$ a scaling factor. Crucially,
adaptation respects the leave-one-incident-out protocol: for each fold, adapters
are trained only on the four training folds and evaluated on the held-out fold,
so no test incident is seen during fine-tuning. A separate adapter is fit per
fold, and adaptation uses the same univariate PM$_{2.5}$ context and the same
per-model input scaling as the corresponding zero-shot variant.

\paragraph{Chronos-2 (LoRA).}
LoRA adapters are fine-tuned for $500$ steps at learning rate $10^{-5}$ with
batch size $256$. Each training example is the concatenation of a window's
48-hour context and its subsequent 24-hour target on the native \ug{} scale, and
the model is optimized with its native forecasting objective. At inference the
adapted model is queried exactly as in the zero-shot case, taking the median
quantile as the point forecast.

\paragraph{Time-MoE (LoRA).}
LoRA adapters of rank $r=8$ (scaling $\alpha=16$, dropout $0.05$) are injected
into the query and value projections of every attention block. The adapter is
trained by next-step prediction with teacher forcing on the per-site standardized
histories---mapping steps $1\!:\!L_{\text{in}}\!-\!1$ to steps
$2\!:\!L_{\text{in}}$---consistent with Time-MoE's autoregressive pretraining
objective. As in the zero-shot case, the horizon is generated autoregressively
and outputs are de-standardized for evaluation.

\subsection{Evaluation Metrics}
\label{sec:metrics}
Forecast accuracy is measured in native units by the mean absolute error and
root-mean-square error over all windows and lead times,
\begin{equation}
\mathrm{MAE}=\frac{1}{Nh}\sum_{i=1}^{N}\sum_{\tau=1}^{h}\big|\hat{y}^{(h)}_{i,\tau}-y^{(h)}_{i,\tau}\big|,
\qquad
\mathrm{RMSE}=\sqrt{\frac{1}{Nh}\sum_{i=1}^{N}\sum_{\tau=1}^{h}\big(\hat{y}^{(h)}_{i,\tau}-y^{(h)}_{i,\tau}\big)^2}.
\label{eq:maermse}
\end{equation}
To complement these absolute-error metrics, we also report the coefficient of
determination,
\begin{equation}
R^{2}=1-\frac{\sum_{i=1}^{N}\sum_{\tau=1}^{h}\big(y^{(h)}_{i,\tau}-\hat{y}^{(h)}_{i,\tau}\big)^{2}}
{\sum_{i=1}^{N}\sum_{\tau=1}^{h}\big(y^{(h)}_{i,\tau}-\bar{y}\big)^{2}},
\label{eq:r2}
\end{equation}
where $\bar{y}$ is the mean of the observed values over all windows and lead
times in the evaluation set. $R^{2}$ measures the fraction of observed variance
explained by the forecasts: it equals one for a perfect forecast, zero for a
forecaster no better than predicting the evaluation-set mean, and becomes
negative when sporadic large-magnitude errors dominate the squared-error sum,
a diagnostic that proves informative for the tail instability analyzed in
Section~\ref{sec:results}.

Because operational value lies in anticipating unhealthy smoke, we additionally
frame each window as an early-warning problem: a window is labeled positive if
its true horizon ever crosses an AQI threshold $t$, and likewise for the
prediction, using the per-window maxima
\begin{equation}
b^{\text{true}}_i=\mathbb{1}\!\left[\max_{\tau} y^{(h)}_{i,\tau} > t\right],
\qquad
b^{\text{pred}}_i=\mathbb{1}\!\left[\max_{\tau}\hat{y}^{(h)}_{i,\tau} > t\right].
\label{eq:exceed}
\end{equation}
From these binary labels we report the exceedance F1-score at the
35.5~\ug{} Unhealthy-for-Sensitive-Groups boundary and the higher AQI
breakpoints (Unhealthy, 55.5; Very Unhealthy, 125.5; Hazardous, 225.5), which
together assess whether a model recovers the rare but consequential extreme
events that dominate the forecasting challenge. All metrics are computed per fold
and averaged across the five leave-one-incident-out folds.

\section{Results}
\label{sec:results}

We evaluate all models under the leave-one-incident-out (LOIO) protocol, reporting
each metric as the mean over five folds and three forecast horizons (6, 12, and
24~hours). Table~\ref{tab:main_results} summarizes overall accuracy and
exceedance-detection performance; Figures~\ref{fig:overall_mae}--\ref{fig:exceedance_f1}
disaggregate these results by horizon and by air-quality category.

\begin{table}[t]
\centering
\caption{Leave-one-incident-out forecasting performance, averaged over five folds and forecast horizons of 6, 12, and 24 hours. MAE and RMSE are in $\mu g\,m^{-3}$ (mean\,$\pm$\,SD across folds); exceedance-detection F1 is reported at the 24-hour PM$_{2.5}$ AQI thresholds for the Unhealthy for Sensitive Groups (35.5), Unhealthy (55.5), Very Unhealthy (125.5), and Hazardous (225.5) categories. Zero-shot foundation models are queried through their public pretrained interfaces without gradient updates; LoRA fine-tuned variants are adapted on each fold's training incidents only, preserving the leave-one-incident-out protocol. Best value per column in \textbf{bold}.}
\label{tab:main_results}
\resizebox{\textwidth}{!}{%
\begin{tabular}{lccccccc}
\toprule
Model & MAE\,$\downarrow$ & RMSE\,$\downarrow$ & $R^2\,\uparrow$ & F1$_{35.5}\uparrow$ & F1$_{55.5}\uparrow$ & F1$_{125.5}\uparrow$ & F1$_{225.5}\uparrow$ \\
\midrule
\multicolumn{8}{l}{\emph{Trained baselines}}\\
Na\"ive Persistence & 6.44\,\small$\pm$0.83 & 16.23\,\small$\pm$2.67 & 0.575 & 0.580 & 0.624 & 0.613 & 0.489 \\
LSTM & 5.28\,\small$\pm$0.59 & 13.22\,\small$\pm$2.17 & 0.722 & 0.643 & 0.697 & 0.692 & 0.565 \\
BiLSTM & \textbf{5.16\,\small$\pm$0.58} & \textbf{12.58\,\small$\pm$2.22} & \textbf{0.746} & \textbf{0.648} & \textbf{0.708} & \textbf{0.714} & \textbf{0.626} \\
Transformer & 5.76\,\small$\pm$0.72 & 13.70\,\small$\pm$2.08 & 0.700 & 0.643 & 0.679 & 0.651 & 0.537 \\
\midrule
\multicolumn{8}{l}{\emph{Foundation models, zero-shot (no gradient updates)}}\\
TimesFM & 5.56\,\small$\pm$0.62 & 14.37\,\small$\pm$2.14 & 0.668 & 0.607 & 0.657 & 0.634 & 0.478 \\
Chronos-2 & 5.61\,\small$\pm$0.69 & 23.45\,\small$\pm$15.34 & -0.075 & 0.640 & 0.688 & 0.666 & 0.537 \\
Moirai-2 & 5.60\,\small$\pm$0.63 & 14.42\,\small$\pm$2.14 & 0.665 & 0.619 & 0.668 & 0.638 & 0.525 \\
Time-MoE & 5.92\,\small$\pm$0.75 & 14.95\,\small$\pm$2.48 & 0.645 & 0.618 & 0.651 & 0.588 & 0.468 \\
\midrule
\multicolumn{8}{l}{\emph{Foundation models, LoRA fine-tuned (per fold)}}\\
Chronos-2 (LoRA-FT) & 5.43\,\small$\pm$0.61 & 15.67\,\small$\pm$4.27 & 0.601 & 0.623 & 0.677 & 0.648 & 0.511 \\
Time-MoE (LoRA-FT) & 5.49\,\small$\pm$0.62 & 14.07\,\small$\pm$2.23 & 0.684 & 0.619 & 0.663 & 0.618 & 0.486 \\
\bottomrule
\end{tabular}%
}
\end{table}

\subsection{Overall Forecasting Accuracy}
\label{sec:results_overall}

\begin{figure}[!htbp]
  \centering
  \includegraphics[width=0.85\linewidth]{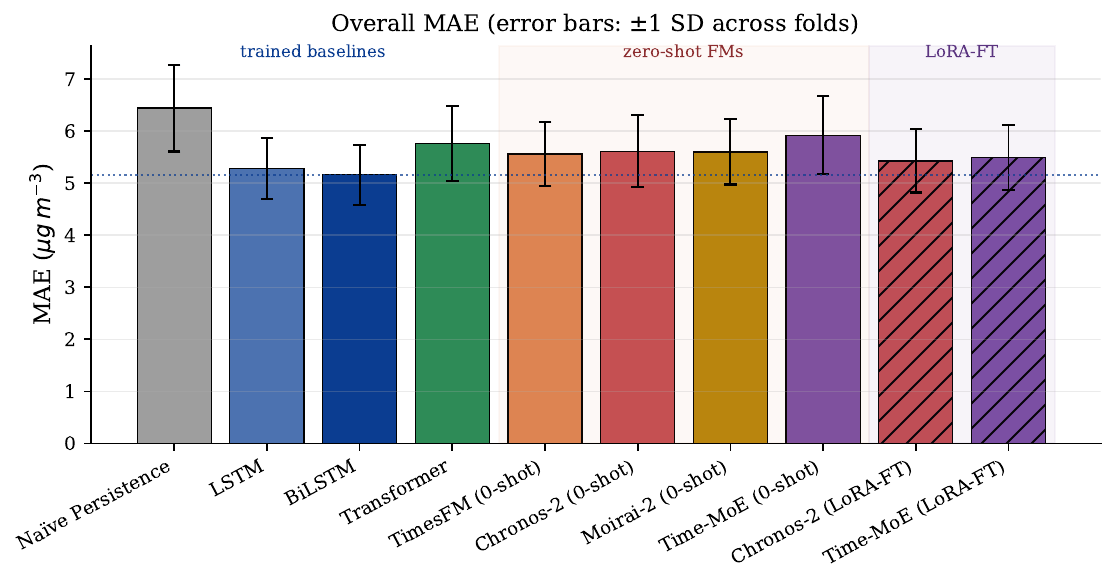}
  \caption{Overall mean absolute error by model, averaged over five folds and
  the 6-, 12-, and 24-hour horizons. Error bars denote $\pm 1$ standard deviation
  across folds; hatched bars mark LoRA fine-tuned foundation-model variants. The
  dotted line marks the best (BiLSTM) MAE. No foundation model matches either
  recurrent baseline.}
  \label{fig:overall_mae}
\end{figure}

The trained bidirectional recurrent baseline attains the best point-forecast
accuracy across every metric. BiLSTM achieves the lowest mean absolute error
(MAE $=5.16\,\mu g\,m^{-3}$) and root mean squared error (RMSE $=12.58\,\mu g\,m^{-3}$)
and the highest coefficient of determination ($R^2=0.75$), followed by the
unidirectional LSTM (MAE $=5.28$, $R^2=0.72$). No zero-shot foundation model
matches either recurrent baseline: the strongest foundation configuration,
Chronos-2 adapted with LoRA fine-tuning, reaches MAE $=5.43\,\mu g\,m^{-3}$, still
above both BiLSTM and LSTM. This ordering is stable across folds, as reflected
in the fold-level standard deviations reported in Table~\ref{tab:main_results}
and visualized in Figure~\ref{fig:overall_mae}.

Two patterns run counter to the common assumption that large pre-trained
transformers dominate sequence tasks. First, the trained Transformer baseline
(MAE $=5.76$, $R^2=0.70$) underperforms both recurrent baselines, indicating
that on a single-site, event-partitioned PM$_{2.5}$ signal the inductive bias of
recurrent encoders is better matched to the task than a from-scratch attention
model. Second, among the foundation models the sparse mixture-of-experts
Time-MoE (MAE $=5.92$ zero-shot) is the weakest, trailing even the na\"ive
persistence floor on the longer horizons, whereas the dense patch-based TimesFM
(MAE $=5.56$) and the universal Moirai-2 (MAE $=5.60$) are more competitive.

\subsection{Error Growth With Forecast Horizon}
\label{sec:results_horizon}

\begin{figure}[!htbp]
  \centering
  \includegraphics[width=\linewidth]{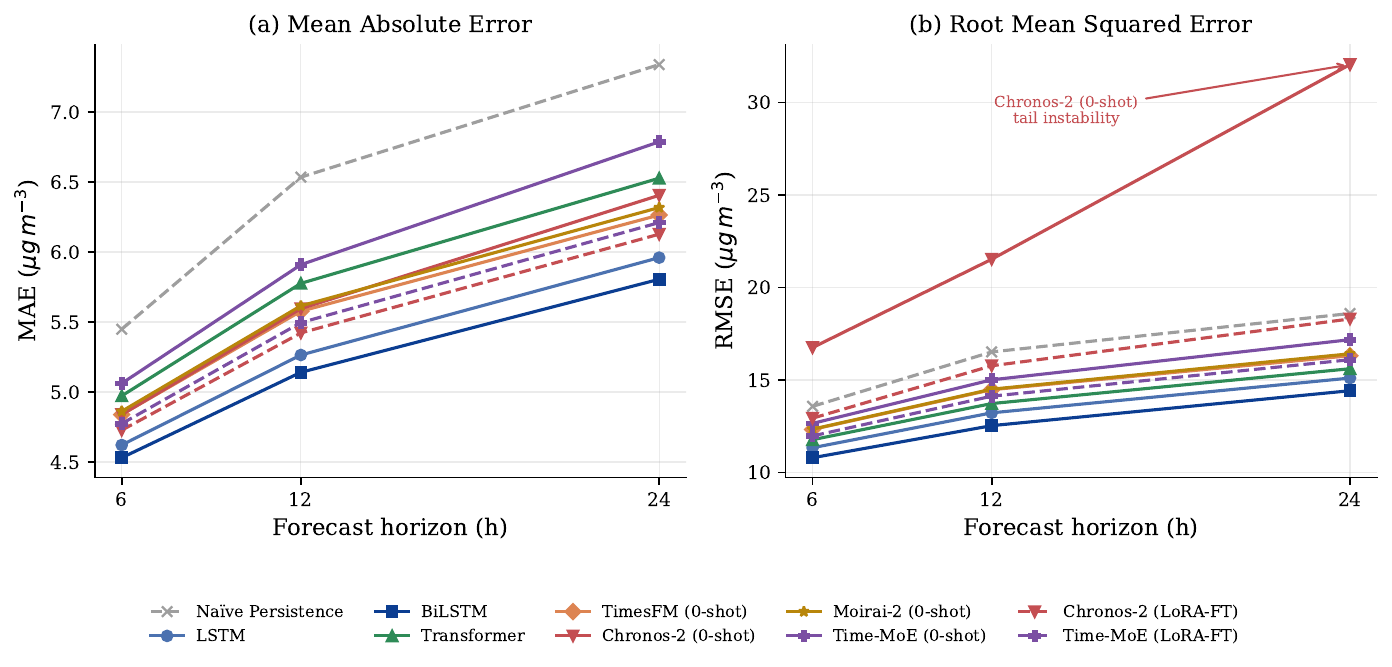}
  \caption{(a) MAE and (b) RMSE as a function of forecast horizon, averaged over
  five folds. Solid lines denote zero-shot foundation models and trained
  baselines; dashed lines denote LoRA fine-tuned foundation-model variants and
  the na\"ive persistence floor. Zero-shot Chronos-2 exhibits a severe RMSE tail
  instability that grows with horizon, despite a competitive MAE.}
  \label{fig:error_vs_horizon}
\end{figure}

All models degrade monotonically as the horizon extends from 6 to 24~hours
(Figure~\ref{fig:error_vs_horizon}). The gap between trained baselines and
foundation models is narrowest at the 6-hour horizon, where several zero-shot
models (TimesFM and Chronos-2 both at MAE $=4.84$) approach BiLSTM
(MAE $=4.53$) and widens with horizon, consistent with foundation models being
most useful at short lead times.

The RMSE panel of Figure~\ref{fig:error_vs_horizon} reveals a pronounced tail
instability for zero-shot Chronos-2. While its MAE remains competitive
($5.61\,\mu g\,m^{-3}$), its RMSE inflates to $23.45\,\mu g\,m^{-3}$ with a large
inter-fold standard deviation ($\pm 15.34$), and its $R^2$ turns negative
($-0.08$ overall, falling to $-0.98$ at the 24-hour horizon). The coexistence of
a moderate MAE with an explosive RMSE indicates that the model produces occasional
large-magnitude errors on a minority of forecasts rather than a uniform loss of
skill, a failure mode that a squared-error criterion penalizes sharply. No other
model exhibits this behavior; the remaining foundation models and all trained
baselines maintain stable RMSE across folds and horizons.

\subsection{Exceedance Detection Across AQI Categories}
\label{sec:results_exceedance}

\begin{figure}[t]
  \centering
  \includegraphics[width=0.78\linewidth]{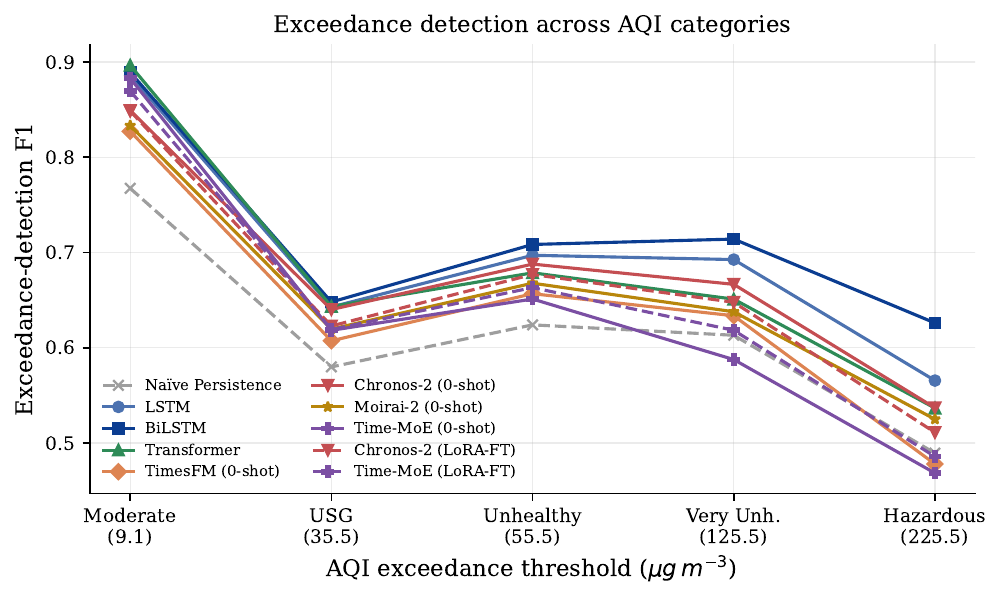}
  \caption{Exceedance-detection F1 across the 24-hour PM$_{2.5}$ AQI category
  thresholds (9.1, 35.5, 55.5, 125.5, and 225.5~$\mu g\,m^{-3}$), averaged over
  five folds and three horizons. All models degrade toward the rarer,
  higher-concentration categories; BiLSTM leads at every threshold, including
  the Hazardous boundary.}
  \label{fig:exceedance_f1}
\end{figure}

As the operational value of wildfire forecasting lies in flagging hazardous
air, we evaluate binary exceedance detection at the 24-hour PM$_{2.5}$ AQI
category thresholds (Figure~\ref{fig:exceedance_f1}). Detection F1 declines for
every model as the threshold rises, reflecting the growing rarity and
burstiness of extreme concentrations. BiLSTM again leads at all thresholds,
including the Hazardous boundary (F1$_{225.5}=0.63$), where the best foundation
model reaches only $0.54$ and the na\"ive baseline falls to $0.49$. The relative
ranking at the high-concentration thresholds mirrors the point-forecast ranking,
indicating that models with lower regression error also transfer better to the
extreme-event regime that matters most for public-health warnings.

\subsection{Effect of LoRA Fine-tuning}
\label{sec:results_finetune}

Parameter-efficient LoRA fine-tuning, applied per fold on the training incidents
only (Section~\ref{sec:fewshot}), improves both foundation families but to
markedly different degrees. For Chronos-2, adaptation reduces RMSE by
$7.78\,\mu g\,m^{-3}$ and raises $R^2$ from $-0.08$ to $0.60$, largely resolving
the tail instability described in Section~\ref{sec:results_horizon} while also
lowering MAE by $0.19\,\mu g\,m^{-3}$. For Time-MoE the gains are smaller but
consistent (MAE $-0.43$, RMSE $-0.88$, $R^2$ $+0.04$), moving it from the
weakest to a mid-ranked foundation model. In both cases, however, the
LoRA-adapted variant remains inferior to the trained BiLSTM on every metric,
reinforcing the central finding that lightweight in-domain adaptation narrows
but does not close the gap to a compact model trained on the target
distribution.

\subsection{Forecast Visualization On A Held-out Incident}
\label{sec:results_showcase}

\begin{figure}[!htbp]
  \centering
  \includegraphics[width=\linewidth]{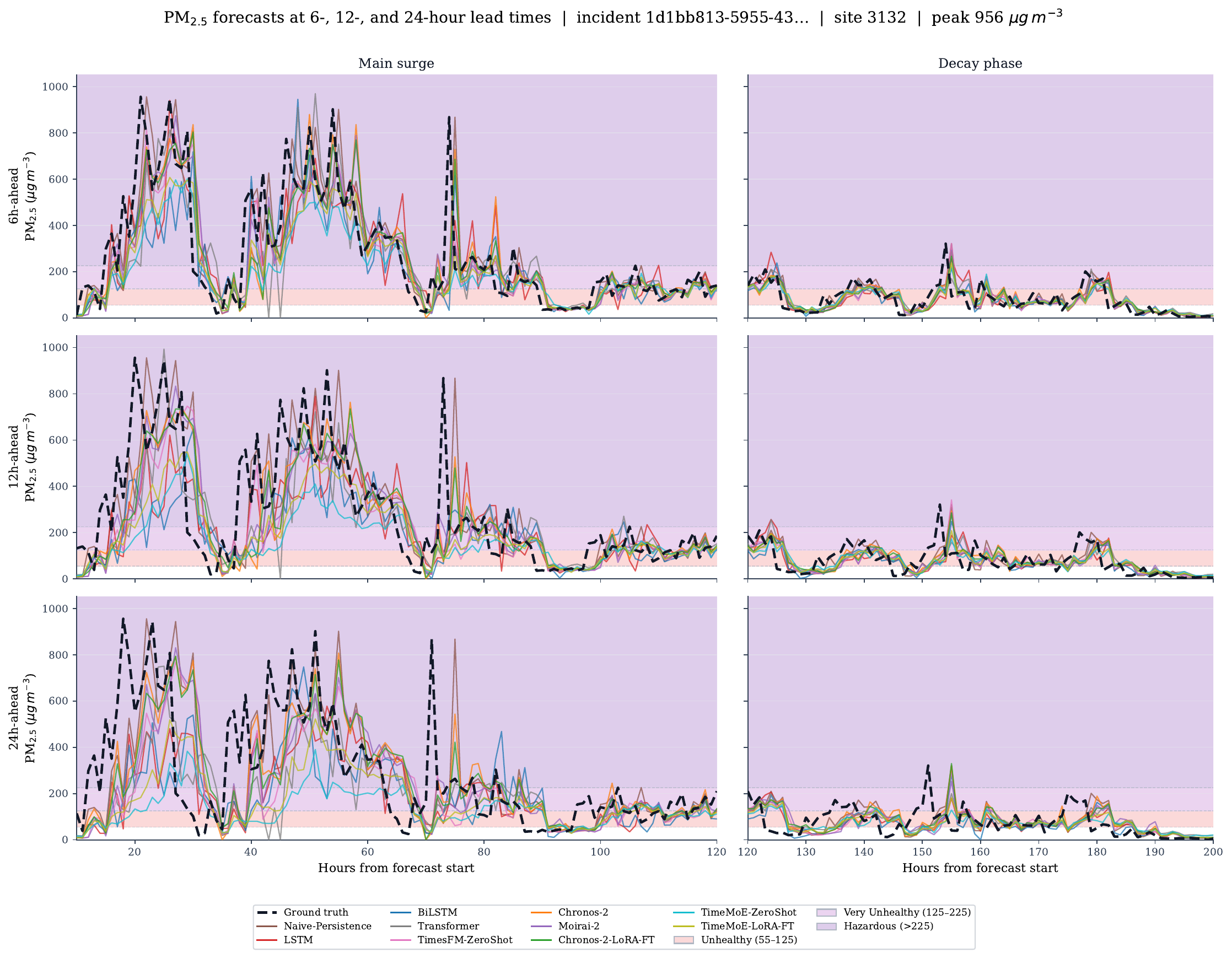}
  \caption{PM$_{2.5}$ forecasts on a held-out severe incident
  (incident~1d1bb813, site~3132) at 6-, 12-, and 24-hour lead times (top to
  bottom), issued at every hour and split into the main surge (10--120\,h) and
  decay phase (120--200\,h). Shaded bands mark the Unhealthy, Very Unhealthy,
  and Hazardous AQI categories; the near-baseline tail beyond 200\,h is
  omitted for clarity. Peak under-prediction and timing misalignment grow with
  lead time.}
  \label{fig:showcase}
\end{figure}

Figure~\ref{fig:showcase} visualizes forecasts on one
held-out test incident (incident~1d1bb813, site~3132), a prolonged severe
episode in which hourly PM$_{2.5}$ repeatedly surged into the Hazardous range,
peaking near $956\,\mu g\,m^{-3}$. For each forecast horizon
$h\in\{6,12,24\}$, we plot the $h$-hour-ahead prediction issued at every hour
of the incident against the corresponding observation, split into the main
surge (10--120~h) and decay phase (120--200~h); PM$_{2.5}$ remains near
baseline for the remainder of the incident window, which is omitted for
clarity.

The visual comparison mirrors the quantitative results in
Table~\ref{tab:main_results}. At the 6-hour lead, all models track both the
timing and, approximately, the amplitude of the concentration surges. Skill
degrades systematically with lead time: at 24~hours the models increasingly
under-predict the sharpest peaks and misalign their timing, and inter-model
divergence widens. Across all horizons, errors concentrate in the excursions
into the Very Unhealthy and Hazardous categories, whereas in the decay phase,
once concentrations recede, trained baselines, zero-shot foundation models,
and LoRA fine-tuned variants become nearly indistinguishable. 

\section{Discussion and Conclusion}
\label{sec:conclusion}

We conducted a systematic benchmark of time series foundation models against
trained deep learning baselines for wildfire PM$_{2.5}$ forecasting under a
leave-one-incident-out evaluation spanning 79 monitoring stations, more than a
decade of hourly observations, ten model configurations, three forecast
horizons, and five AQI severity thresholds. The central result is consistent
across every metric: a compact, fully-trained \model{BiLSTM} achieves the
lowest aggregate error and the highest exceedance F$_1$ at hazardous
concentration levels, and no foundation model, whether zero-shot or LoRA
fine-tuned, matches either recurrent baseline. Zero-shot \model{TSFM}s do improve
on na\"ive persistence in aggregate error, but the margin is modest, and the
gap to trained models widens precisely where it matters most: at longer lead
times and in the extreme concentration regime above the Very Unhealthy and
Hazardous thresholds.

Two mechanisms plausibly explain this ordering. First, the trained baselines
learn exclusively from wildfire-era PM$_{2.5}$ sequences and therefore
internalize the distributional signature of fire smoke, including sharp
onsets, multi-day persistence, and repeated extreme peaks, whereas \model{TSFM}
pre-training corpora contain vanishingly few series in which values exceed
500~\ug{}. Faced with such out-of-distribution magnitudes, foundation models
regress toward typical scales, which simultaneously inflates their error at
the peaks and suppresses the high predictions needed to trigger exceedance
alarms. Second, generic pre-training does not automatically confer stability
in heavy-tailed regimes: zero-shot \model{Chronos-2} pairs a competitive MAE
with an explosive, fold-dependent RMSE and a negative $R^2$, a failure mode
that is invisible to absolute-error metrics and is revealed only by
squared-error and event-level evaluation. This underscores the value of the
\loio{} protocol itself, since chronological splits that leak within-incident
information would have painted a materially more optimistic picture of
\model{TSFM} readiness.

Parameter-efficient adaptation narrows but does not close the gap. LoRA
fine-tuning on each fold's training incidents repairs the \model{Chronos-2}
tail instability and lifts \model{Time-MoE} from the weakest to a mid-ranked
foundation model, yet both adapted variants remain inferior to \model{BiLSTM}
on every metric. The pattern of improvement suggests that lightweight
adaptation primarily recalibrates the magnitude scale of wildfire PM$_{2.5}$
rather than instilling the fine-grained temporal dynamics that govern
exceedance timing. For practitioners, the implication is direct: where a
monitoring site has accumulated multi-year fire history, a small trained
recurrent model remains the strongest and cheapest choice for
exceedance-critical forecasting; where no site-specific data exist, a LoRA
fine-tuned \model{TSFM} offers a credible rapid-deployment path, while unadapted
zero-shot use should be avoided when hazardous-level alarms are the objective.
More broadly, our results indicate that architectural scale and large-scale
pre-training do not substitute for domain-specific training data when the
target includes rare, extreme events.

Several limitations qualify these conclusions and chart the path forward.
All models here receive only the univariate PM$_{2.5}$ channel. A natural next
step is multivariate forecasting that incorporates meteorological covariates
such as wind speed and direction, humidity, and temperature inversions,
together with fire-side drivers such as burned area and fire radiative power,
which recent covariate-capable \model{TSFM} interfaces (e.g., \model{Chronos-2})
now support natively. Equally promising is the integration of spatial
information: our evaluation treats each station independently, whereas smoke
transport couples neighboring monitors, and graph-based or spatiotemporal
architectures that propagate information across the sensor network, or
\model{TSFM}s augmented with spatial context, could improve onset prediction at
stations downwind of an igniting fire. Our protocol also tests incident-level
but not cross-regional generalization; evaluating transfer from California to
other fire-prone regions, and to other extreme air-quality events such as
dust storms, would test whether the conclusions extend beyond a single
airshed. Finally, the adaptation study used a fixed budget of 500 LoRA steps
on 200M-parameter checkpoints, leaving systematic data-efficiency curves,
larger backbones, and physics-informed fine-tuning that embeds transport or
mass-conservation constraints as open directions. As foundation models
continue to scale, the leave-one-incident-out benchmark introduced here
provides a reusable yardstick for measuring whether that progress translates
into reliable warnings for the extreme events that matter most.


\section*{CRediT authorship contribution statement}

\textbf{Yongcan Huang:} Conceptualization, Formal analysis, Investigation,
Writing -- original draft. \textbf{Li Jiang:} Formal analysis, Investigation,
Writing -- original draft. \textbf{Ze Yu Liu:} Conceptualization,
Writing -- review \& editing.

\section*{Data availability}

The hourly PM$_{2.5}$ observations used in this study are publicly available
from the California Air Resources Board Air Quality and Meteorological
Information System (AQMIS) at \url{https://www.arb.ca.gov/aqmis2/aqmis2.php}.
Wildfire incident metadata, including ignition locations, incident dates, and
burned areas, are publicly available from the CAL FIRE incident archive at
\url{https://www.fire.ca.gov/incidents}. 
\bibliographystyle{plainnat}
\bibliography{references}

\end{document}